\documentclass[]{ceurart}

\usepackage{graphicx}
\usepackage[table]{xcolor}

\usepackage{listings}
\usepackage{forest}
\usepackage{adjustbox}
\usepackage{mhchem}

\usepackage{xcolor}

\usepackage{booktabs}
\usepackage{tabularx}
\usepackage{array}
\definecolor{covUniversal}{HTML}{D6E8F5} 
\definecolor{covPartial}{HTML}{FBE7D0}   
\definecolor{covSpecific}{HTML}{ECECEC}  

\usepackage{pifont}

\newcommand{\cmpYes}{\textcolor{green!45!black}{\ding{51}}}
\newcommand{\cmpPart}{\textcolor{orange!85!black}{$\triangle$}}
\newcommand{\cmpNo}{\textcolor{red!70!black}{\ding{55}}}
\newcommand{\cmpNR}{\textcolor{gray}{\textsc{nr}}}

\lstdefinestyle{jsonstyle}{
    basicstyle=\ttfamily\footnotesize,
    frame=single,
    framesep=4pt,
    backgroundcolor=\color{gray!5},
    breaklines=true,
    breakatwhitespace=false,
    showstringspaces=false,
    columns=fullflexible,
    keepspaces=true,
    numbers=none
}

\begin{document}

\copyrightyear{2026}
\copyrightclause{Copyright for this paper by its authors. Use permitted under Creative Commons License Attribution 4.0 International (CC BY 4.0).}

\conference{SeMatS 2026: The 3rd International Workshop on Semantic Materials Science co-located with the 25th International Semantic Web Conference (ISWC 2026), October 25th, 2026, Bari, Italy}

\title{Mined from Scientific Literature: Process Schemas for Atomic Layer Deposition and Etching in Materials Science}

\author[1]{Sameer Sadruddin}[%
orcid=0009-0006-7990-9955,
email=sameer.sadruddin@tib.eu]
\cormark[1]
\address[1]{TIB Leibniz Information Centre for Science and Technology, Hannover 30167, Germany}

\author[2]{Eleni Poupaki}[%
orcid=0000-0001-6607-9813,
email=e.poupaki@tue.nl]
\address[2]{Department of Applied Physics and Science Education, Eindhoven University of Technology, Eindhoven 5600 MB, The Netherlands}

\author[3]{Alex Watkins}[%
orcid=0009-0009-3596-7135,
email=alex.watkins@warwick.ac.uk]
\address[3]{Department of Chemistry, University of Warwick, Coventry CV4 7AL, United Kingdom}

\author[3]{Bora Karasulu}[%
orcid=0000-0001-8129-8010,
email=bora.karasulu@warwick.ac.uk]

\author[2]{Adriaan J. M. Mackus}[%
orcid=0000-0001-6944-9867,
email=A.J.M.Mackus@tue.nl]

\author[2]{Erwin Kessels}[%
orcid=0000-0002-7630-8226,
email=W.M.M.Kessels@tue.nl]

\author[1,4]{Sören Auer}[%
orcid=0000-0002-0698-2864,
email=auer@tib.eu]
\address[4]{L3S Research Center, Leibniz University of Hannover, Hannover 30167, Germany}

\author[1]{Jennifer D'Souza}[%
orcid=0000-0002-6616-9509,
email=jennifer.dsouza@tib.eu]
\cormark[1]

\cortext[1]{Corresponding authors.}

\begin{abstract}
  Atomic layer deposition (ALD) and atomic layer etching (ALE) are reported heterogeneously across experimental and simulation literature in materials science, hindering comparison and machine-actionable reuse. We present four domain-expert-reviewed JSON Schemas for ALD and ALE experimental and simulation processes. Curated with \textsc{schema-miner} and grounded in QUDT using $\textsc{schema-miner}^{pro}$, the schemas structure materials, process conditions, configurations , and measured or predicted results. We compare their scope, structure, and semantic grounding, and demonstrate their use for schema-guided literature extraction and publication of structured records through ORKG templates.
\end{abstract}

\begin{keywords}
atomic layer deposition \sep
atomic layer etching \sep
materials science \sep
semantic models \sep
ontology grounding \sep
knowledge graphs \sep
scientific information extraction
\end{keywords}

\maketitle

\section{Introduction}
\label{sec:introduction}
Atomic layer deposition (ALD) \cite{george2010atomic} and atomic layer etching (ALE) \cite{kanarik2015overview} are complementary cyclic processes for the controlled addition and removal of surface material. ALD deposits thin films through sequential, self-limiting reactions, whereas ALE removes material through self-limiting reaction steps. Their precise control of film thickness, composition, conformality, and material removal makes atomic layer processing important for semiconductor manufacturing and nanoscale materials engineering. Knowledge of these processes is distributed across experimental and computational studies that report different types of information. Experimental studies describe reactants, process conditions, dosing and purging times, growth or etch rates, characterization methods, and measured properties, whereas simulations focus on computational methods, surface models, reaction pathways, and predicted quantities. These perspectives address the same processes but differ substantially in content and organization.

Connecting these perspectives is increasingly important as chemistry and materials research adopts automated experimentation, data-driven analysis, and computational decision-making. Automated thin-film fabrication has already been demonstrated \cite{macleod2020self}, while machine-readable chemical procedures can link published protocols to executable laboratory workflows \cite{mehr2020universal,seifrid2024chemspyd}. Standardized data formats and modular measurement systems further show that interoperable descriptions of experimental inputs, conditions, and outputs are essential for integrating instruments, software, and scientific data \cite{nishio2025digital}. Yet ALD and ALE processes are reported heterogeneously: equivalent concepts use different terminology or units, information is dispersed across text, tables, figures, and supplementary material, and reported parameters and results vary widely. These inconsistencies hinder cross-publication comparison, integration of experimental and simulation results, and reuse in databases, knowledge graphs, information-extraction systems, and machine-learning workflows. Although the AtomicLimits ALD and ALE databases demonstrate the value of structured access \cite{atomiclimits_ald_database,atomiclimits_ale_database}, openly reusable, machine-validatable specifications for both experimental and simulation processes remain unavailable.

In this work, we present four coordinated process schemas: ALD experimental, ALD simulation, ALE experimental, and ALE simulation. Each schema instance represents a single literature-reported process rather than an entire publication, allowing multiple records per study. The schemas organize materials, experimental or computational conditions, process details, and measured or predicted results in nested, process-oriented structures. Separate schemas are required because deposition and etching involve different material transformations, while experiments and simulations have distinct information requirements. The schemas were curated with \textsc{schema-miner} \cite{sadruddin2025llms4schemadiscovery}, a human-in-the-loop workflow combining large language models, representative publications, and iterative domain-expert review. They were then processed with $\textsc{schema-miner}^{pro}$ \cite{sadruddin2025schema} to ground quantitative properties in the Quantities, Units, Dimensions and Data Types (QUDT) ontology \cite{qudt}. The resulting JSON Schemas distinguish required and optional information, constrain datatypes and controlled values, and link quantitative properties to explicit quantity kinds and permitted units. Corresponding ORKG templates support the publication of structured scholarly knowledge \cite{auer2020improving}.

The principal contributions are:

\begin{itemize}
\item four openly available, domain-expert-reviewed JSON Schemas for experimental and simulation processes in ALD and ALE \cite{sadruddin_ald_ale_schemas_2025};

\item a systematic comparison of their scope, structure, required information, domain-property coverage, and QUDT grounding; and

\item a demonstration of schema-guided extraction of ALD process information and publication of structured process records through ORKG templates.
\end{itemize}

Building on our prior \textsc{schema-miner} \cite{sadruddin2025llms4schemadiscovery} and $\textsc{schema-miner}^{pro}$ \cite{sadruddin2025schema} papers, which introduced and empirically evaluated operational workflows for schema discovery and ontology grounding, this paper focuses on the resulting ALD and ALE resources and their application-specific modelling decisions. We examine how these workflows were applied, what the schemas capture, how the four representations relate to one another, and how they enable the machine-actionable reuse of process knowledge.


\section{Related Work}
\label{sec:related-work}

Machine-actionable materials data are represented through complementary resources: ontologies provide shared semantics, domain schemas define specialized record structures, and infrastructures and databases support data management, publication, and retrieval. These resources differ in scope and in whether they describe general knowledge or prescribe individual scientific records. We review them accordingly and position the proposed ALD/ALE schemas as application-level, machine-validatable representations of experimental and simulation processes. 

\begin{table}[!htb]
\centering
\scriptsize
\setlength{\tabcolsep}{2.3pt}
\renewcommand{\arraystretch}{1.12}

\caption{Comparison of representative semantic and machine-actionable
resources relevant to ALD and ALE process representation. Resources are
grouped by their principal representational role.}
\label{tab:related-work-positioning}

\begin{adjustbox}{max width=\textwidth}
\begin{tabular}{@{}p{2.75cm}p{1.10cm}*{12}{c}@{}}
\toprule
\textbf{Resource}
&
\textbf{Type}
&
\multicolumn{4}{c}{\textbf{Scope}}
&
\multicolumn{2}{c}{\textbf{Record Model}}
&
\multicolumn{4}{c}{\textbf{Operationalization}}
&
\multicolumn{2}{c}{\textbf{Construction}}
\\

\cmidrule(lr){3-6}
\cmidrule(lr){7-8}
\cmidrule(lr){9-12}
\cmidrule(lr){13-14}

&
&
\textbf{Exp.}
&
\textbf{Sim.}
&
\textbf{ALD}
&
\textbf{ALE}
&
\begin{tabular}[c]{@{}c@{}}
\textbf{Process}\\
\textbf{record}
\end{tabular}
&
\begin{tabular}[c]{@{}c@{}}
\textbf{Domain}\\
\textbf{detail}
\end{tabular}
&
\textbf{Valid.}
&
\begin{tabular}[c]{@{}c@{}}
\textbf{Req./}\\
\textbf{opt.}
\end{tabular}
&
\begin{tabular}[c]{@{}c@{}}
\textbf{Qty./}\\
\textbf{unit}
\end{tabular}
&
\begin{tabular}[c]{@{}c@{}}
\textbf{Oper.}\\
\textbf{use}
\end{tabular}
&
\begin{tabular}[c]{@{}c@{}}
\textbf{Lit.-}\\
\textbf{inf.}
\end{tabular}
&
\begin{tabular}[c]{@{}c@{}}
\textbf{Domain}\\
\textbf{val.}
\end{tabular}
\\
\midrule

\rowcolor{gray!15}
\multicolumn{14}{l}{
  \textit{Foundational and MSE-wide semantic resources}
}
\\

EMMO \cite{del2024elementary}
& Found.
& \cmpPart & \cmpPart
& \cmpNo & \cmpNo
& \cmpNo & \cmpNo
& \cmpNo & \cmpNo
& \cmpPart & \cmpPart
& \cmpNo & \cmpPart
\\

PMDco \cite{bayerlein2024pmd}
& Mid.
& \cmpYes & \cmpYes
& \cmpNo & \cmpNo
& \cmpPart & \cmpPart
& \cmpPart & \cmpNo
& \cmpPart & \cmpPart
& \cmpNo & \cmpYes
\\

NFDI MatWerk Ont. \cite{beygi2025nfdi}
& RDM
& \cmpPart & \cmpPart
& \cmpNo & \cmpNo
& \cmpNo & \cmpNo
& \cmpNo & \cmpNo
& \cmpNo & \cmpPart
& \cmpNo & \cmpPart
\\

MDS-Onto/MDS \cite{rajamohan2025materials}
& Domain
& \cmpYes & \cmpNo
& \cmpNo & \cmpNo
& \cmpPart & \cmpYes
& \cmpPart & \cmpNo
& \cmpPart & \cmpYes
& \cmpNo & \cmpYes
\\

\rowcolor{gray!15}
\multicolumn{14}{l}{
  \textit{Computational materials resources and infrastructure}
}
\\

ASMO \cite{Azocar_Guzman_Atomistic_Simulation_Methods_2024}
& Domain
& \cmpNo & \cmpYes
& \cmpNo & \cmpNo
& \cmpYes & \cmpYes
& \cmpNo & \cmpNo
& \cmpYes & \cmpNR
& \cmpPart & \cmpNR
\\

NOMAD Metainfo \cite{scheidgen2023nomad}
& Infra.
& \cmpPart & \cmpYes
& \cmpNo & \cmpNo
& \cmpPart & \cmpYes
& \cmpYes & \cmpNR
& \cmpYes & \cmpYes
& \cmpNo & \cmpNR
\\

\rowcolor{gray!15}
\multicolumn{14}{l}{
  \textit{Domain-specific metadata schemas and databases}
}
\\

Plasma-MDS \cite{franke2020plasma}
& Schema
& \cmpYes & \cmpPart
& \cmpNo & \cmpNo
& \cmpPart & \cmpYes
& \cmpYes & \cmpYes
& \cmpNo & \cmpYes
& \cmpYes & \cmpYes
\\

LAMAS 4 IC \cite{wagner2025lamas}
& Schema
& \cmpYes & \cmpNo
& \cmpNo & \cmpNo
& \cmpYes & \cmpYes
& \cmpYes & \cmpYes
& \cmpPart & \cmpYes
& \cmpYes & \cmpYes
\\

AtomicLimits ALD/ALE databases
\cite{atomiclimits_ald_database,atomiclimits_ale_database}
& DB
& \cmpYes & \cmpNo
& \cmpYes & \cmpYes
& \cmpPart & \cmpPart
& \cmpNo & \cmpNR
& \cmpNo & \cmpYes
& \cmpYes & \cmpYes
\\

\rowcolor{gray!22}
\multicolumn{14}{l}{
  \textit{Proposed resource}
}
\\

\textbf{ALD/ALE process schemas}
& \textbf{Schema}
& \cmpYes & \cmpYes
& \cmpYes & \cmpYes
& \cmpYes & \cmpYes
& \cmpYes & \cmpYes
& \cmpYes & \cmpYes
& \cmpYes & \cmpYes
\\

\bottomrule
\end{tabular}
\end{adjustbox}

\vspace{2pt}

\begin{minipage}{\textwidth}
\scriptsize
\textit{Note:}
\cmpYes{} denotes explicit support;
\cmpPart{} denotes generic, indirect, or partial support;
\cmpNo{} denotes that the capability is not provided; and
\cmpNR{} denotes information not reported or not verifiable.
ALD and ALE are marked only when represented as explicit application
domains rather than through generic extensibility.
``Exp.'' and ``Sim.'' denote explicit representation of experimental
and computational processes, respectively.
``Process record'' denotes a structure that can represent one individual
scientific process.
``Domain detail'' denotes representation of domain-specific inputs,
materials or reactants, conditions, equipment or model parameters, and
outputs.
``Valid.'' denotes machine validation of populated records; OWL
reasoning alone is not counted.
``Req./opt.'' denotes an explicit distinction between mandatory and
optional reporting fields.
``Qty./unit'' denotes explicit quantity-kind or unit representation or
constraints.
``Oper. use'' denotes a publicly available implementation, populated
resource, data-entry mechanism, or demonstrated downstream application.
``Lit.-inf.'' denotes that scientific literature was directly used to
derive or refine the representation.
``Domain val.'' denotes reported domain-expert, community, or laboratory
validation.
Type abbreviations:
Found.\ = foundational ontology;
Mid.\ = mid-level ontology;
RDM = research-data-management ontology;
Infra.\ = infrastructure schema;
DB = database.
For a broader collection of ontologies and semantic resources for materials science and engineering, see the OntoLearner library\footnote{\url{https://ontolearner.readthedocs.io/benchmarking/benchmark.html\#materials-science-and-engineering}} \cite{giglou2026ontolearner,ontolearner-software}.
\end{minipage}
\end{table}

\subsection{Semantic resources for materials data and simulation}

The Elementary Multiperspective Material Ontology (EMMO) \cite{del2024elementary} provides a physics-based foundation for representing materials, physical entities, properties, modelling, and characterisation, while the Platform MaterialDigital Core Ontology (PMDco) \cite{bayerlein2024pmd} adds mid-level concepts for materials, manufacturing processes, experiments, measurements, simulations, and data transformations. The NFDI MatWerk Ontology \cite{beygi2025nfdi} supports the description and discovery of research resources such as projects, datasets, workflows, software, instruments, facilities, services, and metadata schemas. MDS-Onto \cite{rajamohan2025materials} provides a modular framework for developing BFO-aligned materials and data-science ontologies and publishing interoperable JSON-LD data through the FAIRmaterials and FAIRLinked toolchains. Together, these resources support semantic interoperability, data organization, and publication, but they do not prescribe machine-validatable records for individual ALD or ALE processes or specify how process-specific information such as precursor doses, purge durations, supercycles, growth or etch per cycle, and process windows should be organized.

For computational materials research, the Atomistic Simulation Methods Ontology (ASMO) \cite{Azocar_Guzman_Atomistic_Simulation_Methods_2024} represents methods such as density functional theory, molecular dynamics, and Monte Carlo simulation, together with their inputs, parameters, and outputs. However, it does not connect these methods to ALD/ALE-specific precursors, surfaces, process conditions, reaction behaviour, or resulting material properties. NOMAD \cite{scheidgen2023nomad} addresses computational materials data at the infrastructure level through its extensible Metainfo system, which organizes descriptions, datatypes, units, shapes, and allowed values into hierarchical sections, subsections, references, and quantities. While this supports normalization across simulation codes and materials domains, the proposed schemas serve a narrower, complementary role by defining compact, code-independent records for individual ALD and ALE simulations.

\subsection{Domain-specific schemas and structured ALD/ALE resources}

Domain-specific metadata schemas translate broad semantic models into machine-validatable records for scientific procedures. Plasma-MDS \cite{franke2020plasma} standardizes plasma-science research data through concepts such as plasma source, medium, target, simulation method, and associated digital resources, and uses JSON Schema to support validation and integration into data-management systems. LAMAS 4 IC \cite{wagner2025lamas} similarly provides ASTM E1151-aligned JSON Schemas for anion and cation-exchange chromatography, combining shared modelling principles with procedure-specific structures, controlled terminology, laboratory validation, and explicit mandatory and optional fields. These resources provide important methodological precedents, but neither covers ALD or ALE or jointly represents experimental and simulation processes.

The closest ALD/ALE resources are the community-curated AtomicLimits databases \cite{atomiclimits_ald_database,atomiclimits_ale_database}, which organize literature records by deposited or etched materials, reactants, and publications and assign persistent identifiers for reuse. However, they primarily support curation and access rather than validation of independently created records and do not constrain quantity kinds or units. More broadly, the reviewed resources provide semantic foundations, research-data models, simulation infrastructures, domain schemas, or ALD/ALE databases, but none combines both ALD and ALE, separate experimental and simulation models, process-level records, closed-world validation, required and optional fields, and ontology-grounded quantities and units. The proposed resource addresses this gap through four coordinated, literature-derived, domain-expert-reviewed, and QUDT-grounded schemas.

\section{ALD/E Experimental/Simulation Schema Curation Methodology}
\label{sec:schema-curation-methodology}

The four schemas presented in this work were curated using \textsc{schema-miner} \cite{sadruddin2025llms4schemadiscovery} which was applied independently to the experimental ALD, simulation ALD, experimental ALE, and simulation ALE studies. Experimental and simulation studies report structurally different information: experimental publications describe measured process conditions (e.g., precursor dosing and purge times, substrate temperature) and characterized material, whereas simulation studies report computational method parameters (e.g., the simulation method used together with method-specific settings such as DFT functionals) and predicted or modeled quantities. This distinction motivated curating separate schemas for the experimental and simulation perspectives of each process, rather than a single merged representation.

\subsection{\textsc{schema-miner}}
The \textsc{schema-miner}\cite{sadruddin2025llms4schemadiscovery} workflow consists of three stages. In the first stage, an initial schema is generated from a domain-specific process specification document that describes the major entities, parameters, and outputs expected for the process. The second stage performs literature-guided refinement using a relatively small set of representative scientific publications. Information reported in these publications is compared against the initial schema to identify missing properties, insufficiently specified concepts, inappropriate datatypes, and structural inconsistencies. The schema is subsequently revised using both the literature-derived evidence and feedback supplied by domain experts. In the third and final stage, the refined schema is applied to a substantially larger publication corpus. This stage broadens the coverage of the schema by exposing it to a wider range of materials systems, process configurations, and experimental conditions. The iterative process continues until the domain-experts consider the resulting schema sufficiently complete, correct, and representative of the corresponding ALD or ALE literature.

The publication corpora used during this work were curated by domain experts in the ALD and ALE domain. The main selection criterion was scientific quality, while also ensuring coverage of diverse materials, process conditions, methodological scenarios, and application areas within ALD and ALE. The objective was to expose \textsc{schema-miner} to high-quality studies covering a broad range of relevant reporting practices. During the refinement stage (Stage 2), 7 publications were used for each of the ALD experimental and simulation schemas, 9 for the ALE experimental schema, and 10 for the ALE simulation schema. During the finalization stage (Stage 3), the corpora were expanded to 59 publications for ALD experimental, 51 for ALD simulation, 38 for ALE experimental, and 22 for ALE simulation.

\subsection{Ontology Grounding with $\textsc{schema-miner}^{pro}$}
Following the structural curation, the schemas were processed using $\textsc{schema-miner}^{pro}$ \cite{sadruddin2025schema} to semantically ground the properties representing a physical quantity in the QUDT ontology \cite{qudt}. The grounding provides a common semantic reference for the meaning of physical quantities and their permitted units. It supports the normalization of heterogeneous units encountered in the literature, and facilitates machine-assisted comparison of values across publications. Each grounded property contains a numeric value, a QUDT quantity-kind IRI, and one or more QUDT unit IRIs. In the JSON Schemas, these mappings are represented through the \texttt{quantityValue}, \texttt{quantityKind}, \texttt{hasQuantityKind}, and \texttt{sameAs} fields. Table~\ref{tab:overview-stats} reports the number of QUDT-grounded properties in each final schema.

The final schemas are openly accessible on Zenodo \cite{sadruddin_ald_ale_schemas_2025}. In addition to the JSON Schema, all four schemas were also translated into ORKG templates\footnote{ALD experimental: \url{https://orkg.org/templates/R1434174}; ALD simulation: \url{https://orkg.org/templates/R1368294}; ALE experimental: \url{https://orkg.org/templates/R1379646}; ALE simulation: \url{https://orkg.org/templates/R1434008}.} to support structured scholarly knowledge representation within the Open Research Knowledge Graph \cite{auer2020improving}. Section~\ref{sec:schema-specification} describes the structure of the four resulting schemas in detail, including the design decisions specific to each of the experimental and simulation perspectives introduced above.

\section{Schema Specifications}
\label{sec:schema-specification}

This section specifies the four schemas in detail. We first describe their shared scope and the modeling principles common to all four (Sections~\ref{sec:overview} and~\ref{sec:common-principles}), then present each schema individually (Sections~\ref{sec:ald-exp}--\ref{sec:ale-sim}), and finally compare them directly to show where they converge and diverge (Section~\ref{sec:cross-schema}).

\subsection{Overview and Scope}
\label{sec:overview}

The four schemas presented in this work formalize how ALD and ALE processes are described in the scientific literature. Each schema instance represents a single experimental or simulation process record rather than an entire publication, where a publication can report several distinct processes, for example, the same chemistry studied at different deposition temperatures, or with different precursors. Each record captures the process condition, the materials involved, the relevant experimental or simulation conditions, and the resulting measured or predicted properties. Collectively, the schemas are designed to support questions such as: \textit{What growth conditions produced this ALD film?} \textit{What etch chemistry and process parameters were used to remove this material?} and \textit{What computational method and settings produced this simulated ALD outcome?}

Table~\ref{tab:overview-stats} summarizes the structural characteristics of the four schemas. The schemas are organized as nested, process-oriented JSON Schema structures but differ substantially in their number of declared properties, nesting depth, and required properties. These differences reflect the distinct structures and minimum reporting requirements identified for experimental and simulation studies of ALD and ALE.

\begin{table}[!t]
\centering
\caption{Structural and QUDT-grounding statistics for the four ALD/ALE experimental and simulation JSON Schemas, including maximum nesting depth, property counts with and without QUDT helper fields, QUDT-grounded properties, distinct quantity kinds and units, and required domain properties.}
\label{tab:overview-stats}
\resizebox{\textwidth}{!}{%
\begin{tabular}{lcccccccc}
\hline
\multicolumn{1}{c}{\textbf{Schema}} & \textbf{\begin{tabular}[c]{@{}c@{}}Top-level \\ properties\end{tabular}} & \textbf{\begin{tabular}[c]{@{}c@{}}Max \\ depth\end{tabular}} & \textbf{\begin{tabular}[c]{@{}c@{}}Required \\ properties\end{tabular}} & \textbf{\begin{tabular}[c]{@{}c@{}}Total \\ properties\\ (incl. \\ QUDT)\end{tabular}} & \textbf{\begin{tabular}[c]{@{}c@{}}Total \\ properties \\ (excl. \\ QUDT)\end{tabular}} & \textbf{\begin{tabular}[c]{@{}c@{}}Total \\ QUDT-grounded \\ properties\end{tabular}} & \textbf{\begin{tabular}[c]{@{}c@{}}Distinct \\ QUDT \\ kinds\end{tabular}} & \textbf{\begin{tabular}[c]{@{}c@{}}Distinct \\ QUDT \\ units\end{tabular}} \\ \hline
\begin{tabular}[c]{@{}l@{}}ALD \\ Experimental\end{tabular} & 6 & 8 & 30 & 275 & 95 & 30 & 18 & 33 \\
\begin{tabular}[c]{@{}l@{}}ALD \\ Simulation\end{tabular} & 8 & 6 & 3 & 212 & 68 & 24 & 15 & 26 \\
\begin{tabular}[c]{@{}l@{}}ALE \\ Experimental\end{tabular} & 11 & 7 & 20 & 167 & 47 & 20 & 14 & 20 \\
\begin{tabular}[c]{@{}l@{}}ALE \\ Simulation\end{tabular} & 8 & 7 & 56 & 275 & 83 & 32 & 14 & 23 \\ \hline
\end{tabular}%
}
\end{table}

\subsection{Common Modeling Principles}
\label{sec:common-principles}

Although the four schemas differ in scope and content, they share a common set of modeling conventions, applied consistently to keep the schemas mutually comparable and to support their reuse in the community. Each schema organizes properties into top-level groups reflecting stages of the underlying process (e.g., process parameters, reactant selection, resulting material properties). Table~\ref{tab:overview-stats} reports the number of top-level groups and the maximum nesting depth for each schema. In addition, every property is given a descriptive, camel-case identifier (e.g., \texttt{plasmaPower}), and is accompanied by a natural-language \texttt{description} field clarifying its meaning, expected use, or applicable conditions, for instance, the \texttt{plasmaPower} property is annotated as ``Plasma power (in Watts) if PEALD was used.''.

Each schema explicitly distinguishes properties that must be present in a valid instance from those that may be omitted, using the JSON Schema \texttt{required} keyword. Numeric properties representing physical quantities are not encoded as numbers. Instead, they are represented as nested objects containing a \texttt{quantityValue} field, holding a \texttt{numericValue} and an associated \texttt{unit}, separating the magnitude of a measurement from its unit and avoiding ambiguity when the same quantity is reported in different units across the literature (e.g., plasma power reported in watts and milliwatts in different publications).

Beyond separating value from unit, each physical quantity is further grounded in the QUDT ontology (Section~\ref{sec:schema-curation-methodology}): a \texttt{quantityKind} field records the QUDT quantity-kind IRI (e.g., \url{https://qudt.org/vocab/quantitykind/Power}), and the associated \texttt{unit} object constrains \texttt{hasQuantityKind} and \texttt{sameAs} to QUDT-defined quantity-kind and unit IRIs, respectively. Figure~\ref{lst:plasmapower} in Appendix, shows this grounding for the \texttt{plasmaPower} property of the ALD-experimental schema, which enables the normalization and machine-assisted comparison of quantities reported under heterogeneous unit conventions.

\subsection{ALD Experimental Schema}
\label{sec:ald-exp}

The ALD experimental schema contains the largest number of domain-properties and is the most deeply nested of the four schemas. It comprises 275 properties organised into six top-level groups and reaching a maximum nesting depth of eight levels. Its structure is shown in Figure~\ref{fig:ald-exp-tree}. The schema captures a complete experimental deposition record, from the chemistry and process conditions through to the resulting film and the performance of devices fabricated from that film. The six top-level groups partition the record along familiar experimental lines. \texttt{aldSystem} identifies the deposition method and the material deposited while \texttt{reactantSelection} records the precursors, co-reactants, and process gases and \texttt{processParameters} holds the reactor conditions, thickness control, and supercycle design. The outcome of the process is described by \texttt{materialProperties}, which covers optical, electrical, compositional, and morphological film characteristics, while \texttt{deviceProperties} captures the electrical behaviour of devices using this film. A final \texttt{otherAspects} group records safety, film stability, and reproducibility.

The schema's unique design decision is its compound-centric representation of process chemistry. The \texttt{aldMethod}, \texttt{precursor}, and \texttt{coReactant} properties are each modelled as arrays of objects that pair a named compound with its associated role, so that a single record can describe a multi-component process, such as the deposition of a ternary material in which different sub-compounds are deposited by different reactants. The same philosophy recurs at finer granularity in \texttt{filmComposition}, where elemental concentration is represented as an array of {element, concentration} pairs. This allows compositions of multicomponent materials, for example, the In:Ga:Zn metal ratio in an InGaZnO film, to be recorded per element rather than compressed into a single value. The \texttt{supercycleDesign} group complements this by capturing the cyclic ordering of subcycles in multi-material processes: a supercycle defines a repeating sequence of subcycles, each a standard ALD cycle for one material, combined in a specified ratio to build up ternary, quaternary compounds, nanolaminates, or doped films with controlled composition. Another distinguishing feature is the \texttt{deviceProperties} group, which is unique to this schema among the four and reflects downstream device performance, in which the specific film has been integrated, recording quantities such as field-effect mobility, threshold voltage, subthreshold swing, and on/off ratio. Evaluating these quantities is an important part of assessing ALD film quality for its intended industrial application, and the relevant properties can vary by material and device type; in our case, we selected thin-film transistors (TFTs) as the target application.

All the six top-level groups are required. A record must identify its chemistry where the compound and its associated precursor or co-reactant are mandatory within each reactant entry. Most quantitative process and material properties are optional, reflecting that curated literature reports conditions and measurements to varying degrees. The schema grounds 30 quantitative properties in QUDT, spanning 18 distinct quantity kinds and 33 distinct units. The ALD simulation schema which is presented next, shares the ALD process description from a simulation perspective.

\begin{figure}[!htbp]
\centering
\begin{adjustbox}{max width=\textwidth}
\begin{forest}
for tree={
  grow'=0,
  parent anchor=east,
  child anchor=west,
  anchor=west,
  edge path={
    \noexpand\path[\forestoption{edge}]
      (!u.parent anchor) -- +(5pt,0) |- (.child anchor)\forestoption{edge label};
  },
  edge={thin},
  l sep=10pt,
  s sep=2pt,
  font=\footnotesize,
}
[ALD experimental schema
  [aldSystem$^{*}$
    [{aldMethod$^{*}$ : array<object>}
      [{compound$^{*}$ : string}]
      [{method$^{*}$ :
        enum\{PEALD, TALD, SALD\}}]
    ]
    [materialDeposited$^{*}$ : string]
  ]
  [reactantSelection$^{*}$
    [precursor$^{*}$ : array<object>
      [compound$^{*}$ : string]
      [precursor$^{*}$ : string]
    ]
    [coReactant$^{*}$ : array<object>
      [compound$^{*}$ : string]
      [coReactant$^{*}$ : string]
    ]
    [carrierGas : string]
    [purgingGas : string]
  ]
  [processParameters$^{*}$
    [reactor : object$^\ddagger$]
    [plasmaPower
      [quantityValue
        [numericValue : number]
        [unit]
      ]
      [quantityKind]
    ]
    [substrate : string]
    [deliveryMethod : string]
    [temperature : QUDT quantity$^\dagger$]
    [pressure : QUDT quantity$^\dagger$]
    [thicknessControl : object$^\ddagger$]
    [supercycleDesign : object$^\ddagger$]
  ]
  [materialProperties$^{*}$
    [opticalProperties : object$^\ddagger$]
    [electricalProperties : object$^\ddagger$]
    [uniformity : object$^\ddagger$]
    [conformality : object$^\ddagger$]
    [filmComposition : object$^\ddagger$]
    [chemicalComposition : string]
    [crystallinity : object$^\ddagger$]
    [roughness : object$^\ddagger$]
    [filmDensity : object$^\ddagger$]
  ]
  [deviceProperties$^{*}$
    [fieldEffectMobility : QUDT quantity$^\dagger$]
    [thresholdVoltage : QUDT quantity$^\dagger$]
    [subthresholdSwing : QUDT quantity$^\dagger$]
    [onOffRatio : QUDT quantity$^\dagger$]
    [deviceStructure : string]
  ]
  [otherAspects$^{*}$
    [safety : string]
    [filmStability : boolean]
    [reproducibility : boolean]
  ]
]
\end{forest}
\end{adjustbox}

\vspace{2pt}
{\footnotesize
$^{*}$ Required within its containing object;\quad
$^{\dagger}$ QUDT-grounded quantitative property;\quad
$^{\ddagger}$ additional nested structure omitted.
}

\caption{Top-level structure of the ALD-experimental schema. Most leaf nodes shown are the immediate children of each top-level property group. The \texttt{precursor}/\texttt{coReactant} branch is expanded to illustrate per-compound reactant modeling, and \texttt{plasmaPower} is expanded to illustrate the \texttt{quantityValue}/\texttt{quantityKind} representation of QUDT-grounded quantities.}
\label{fig:ald-exp-tree}
\end{figure}

\subsection{ALD Simulation Schema}
\label{sec:ald-sim}

The ALD simulation schema describes computational studies of the atomic layer deposition process, organised into eight top-level groups across 212 properties and a maximum depth of six levels. Its structure is shown in Figure~\ref{fig:ald-sim-tree}. The eight top-level groups capture the simulation perspective of an ALD process. \texttt{simulationParameters} captures the computational method and its settings, \texttt{materials} identifies the modeled chemistry, and \texttt{reactorConditions} holds the simulated process environment. The predicted behaviour of the process is distributed across \texttt{growthRate}, \texttt{surfaceProperties}, \texttt{filmProperties}, \texttt{nucleationProcess}, and \texttt{selectiveGrowthMechanism}, which together capture growth kinetics, surface reactions, resulting film characteristics, initial nucleation behaviour, and the mechanisms underlying area-selective deposition.

One design decision distinguishes this schema from its experimental counterpart: a dedicated treatment of surface and mechanistic detail. \texttt{surfaceProperties} and \texttt{nucleationProcess} capture quantities such as sticking coefficients, chemisorbed precursor densities, surface hydroxyl concentrations, and nucleation delay, which are natural outputs of a simulation but are rarely reported directly in experimental work. Of the eight top-level groups, only \texttt{simulationParameters} and \texttt{materials} are mandatory, and within \texttt{simulationParameters} only the \texttt{methods} field is required. This reflects the wide variation in what computational studies report: a record must identify its method and chemistry, but little else is universally present. The schema grounds 24 quantitative properties in QUDT across 15 distinct quantity kinds and 26 distinct units. The two ALD schemas together describe deposition from complementary experimental and computational viewpoints; Section~\ref{sec:ale-exp} turns to the etching counterpart, where the same experiment–simulation division recurs over a distinct process.

\begin{figure}[!htbp]
\centering
\begin{adjustbox}{max width=\textwidth}
\begin{forest}
for tree={
  grow'=0,
  parent anchor=east,
  child anchor=west,
  anchor=west,
  edge path={
    \noexpand\path[\forestoption{edge}]
      (!u.parent anchor) -- +(5pt,0) |- (.child anchor)\forestoption{edge label};
  },
  edge={thin},
  l sep=10pt,
  s sep=1pt,
  font=\footnotesize,
}
[ALD simulation schema
  [simulationParameters$^{*}$
    [methods : array<enum>]
    [methodDetails : object$^\ddagger$]
    [source : object$^\ddagger$]
  ]
  [materials$^{*}$
    [precursors : array<string>]
    [coReactants : array<string>]
    [substrates : array<string>]
    [encapsulationMaterials : array<string>]
    [ligandModification : object$^\ddagger$]
  ]
  [growthRate
    [rate : QUDT quantity$^\dagger$]
    [temperatureDependence : QUDT quantity$^\dagger$]
    [propertySource : string]
  ]
  [surfaceProperties
    [desorptionRate : QUDT quantity$^\dagger$]
    [diffusionRate : QUDT quantity$^\dagger$]
    [reactionRate : QUDT quantity$^\dagger$]
    [stickingCoefficient : QUDT quantity$^\dagger$]
    [bindingAffinity : QUDT quantity$^\dagger$]
    [surfaceCoverage : object$^\ddagger$]
    [chemisorptionCharacteristics : object$^\ddagger$]
    [reactionPathways : array<object>$^\ddagger$]
    [surfaceTerminationChemistry : string]
    [propertySource : string]
  ]
  [filmProperties
    [uniformity : QUDT quantity$^\dagger$]
    [roughness : QUDT quantity$^\dagger$]
    [density : QUDT quantity$^\dagger$]
    [temperatureProfile : array<object>$^\ddagger$]
    [chemicalComposition : array<string>]
    [propertySource : string]
  ]
  [reactorConditions
    [pressure : QUDT quantity$^\dagger$]
    [carrierGasFlow : QUDT quantity$^\dagger$]
    [carrierGasType : string]
    [precursorFlow : object$^\ddagger$]
    [gapDistance : QUDT quantity$^\dagger$]
    [propertySource : string]
  ]
  [selectiveGrowthMechanism
    [facetPreference : string]
    [substituentEffects : string]
    [blockingMechanisms : object$^\ddagger$]
  ]
  [nucleationProcess
    [nucleationDelay : QUDT quantity$^\dagger$]
    [selfCleaningEffect : boolean]
  ]
]
\end{forest}
\end{adjustbox}

\vspace{2pt}
{\footnotesize $^{*}$ Required within its containing object;\quad $^\dagger$ QUDT-grounded quantitative property;\quad $^\ddagger$ additional nested structure omitted.}

\caption{Top-level structure of the ALD-simulation schema. Leaf nodes are the immediate children of each top-level property group; nested structure follows the conventions illustrated in Figure~\ref{fig:ald-exp-tree}.}
\label{fig:ald-sim-tree}
\end{figure}

\subsection{ALE Experimental Schema}
\label{sec:ale-exp}

The ALE experimental schema records experimental studies of atomic layer etching. It comprises 167 properties organised into eleven top-level entries, the largest number of top-level entries among the four schemas. Its structure is shown in Appendix: Figure~\ref{fig:ale-exp-tree}. The eleven top-level properties describe the etch process and its outcome. Three top-level properties provide a concise identification of the process: \texttt{etchedMaterial} specifies the target material, \texttt{processType} identifies the ALE process, and \texttt{directionality} distinguishes anisotropic from isotropic etching. The remaining properties are organized into eight object groups. \texttt{reactantSelection} records the reactants used in the alternating surface-modification and material-removal steps, while \texttt{processDetails} captures the substrate, reactor, operating conditions, pulse and purge durations, reactant flow, and number of cycles. \texttt{etchControl} records etch-per-cycle and mass-loss measurements, whereas \texttt{synergy} quantifies the extent to which material removal results from the combined action of the two half-reactions. \texttt{aleWindow} describes the temperature and ion-energy ranges over which self-limiting etching is maintained. The resulting material and process outcomes are represented through \texttt{etchedMaterialProperties}, \texttt{selectivity}, and \texttt{otherAspects}, which cover surface and material characteristics, preferential etching relative to other exposed materials, and safety, sustainability, and environmental considerations.

All eleven top-level properties are required, so every record must at minimum identify its material, process type, directionality, reactants, and each of the process and outcome groups. Within those groups, individual quantitative properties remain largely optional. The schema grounds 20 quantitative properties in QUDT across 14 distinct quantity kinds and 20 distinct units. The ALE simulation schema, described next, complements this experimental view with the predicted quantities and surface-level mechanisms of computational etching studies.

\subsection{ALE Simulation Schema}
\label{sec:ale-sim}

The ALE simulation schema describes computational studies of atomic layer etching. It comprises 275 properties organised into eight top-level groups, reaches a maximum nesting depth of seven levels. Its structure is shown in Appendix: Figure~\ref{fig:ale-sim-tree}. As with the ALD simulation schema, it records what an etching simulation predicts and the methodology behind it. The eight top-level groups combine process specification with predicted results. \texttt{processType}, \texttt{chemicalPrecursors}, \texttt{energySource}, and \texttt{substrateMaterial} identify the etch process and the system being modelled. \texttt{surfaceModel} describes the atomistic surface used in the simulation and \texttt{processParameters} holds the simulated conditions. The outcomes are gathered under \texttt{simulationResults}, which covers the etched material, by-products, surface modifications, decomposition mechanisms, and predicted rates and properties, while \texttt{simulationMethodology} records the computational methods employed.

The breadth of \texttt{simulationResults} is a distinguishing feature, encompassing not only predicted quantities but also qualitative mechanistic outcomes such as decomposition and self-limiting-reaction behaviour. The ALE simulation schema requires all eight top-level properties, together with a substantial number of properties within them, making it the most prescriptive of the four schemas. The schema grounds 32 quantitative properties in QUDT, the most of any of the four schemas and across 14 distinct quantity kinds and 23 distinct units, constraining 29 of them to physically meaningful numeric ranges. Now having presented the four schemas individually, we now turn to their relationships: Section~\ref{sec:cross-schema} compares them directly, examining where they share a common core and where each diverges to serve its particular process and perspective.

\begin{figure}[!t]
    \centering
    \includegraphics[width=1\textwidth]{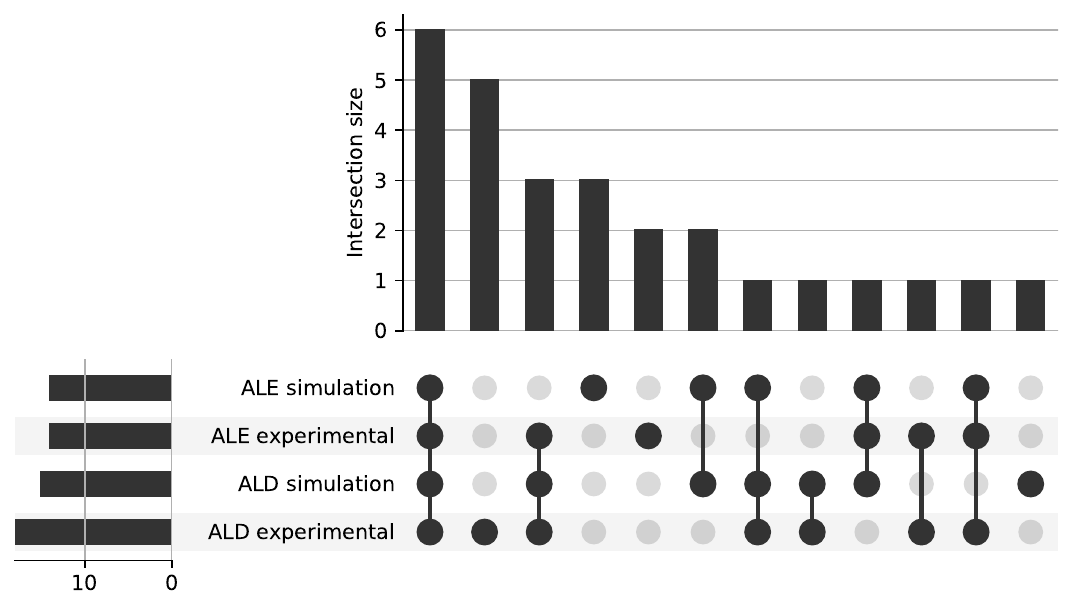}
    \caption{UpSet plot of the QUDT quantity kinds shared across the four ALD/ALE schemas. Each matrix column is an exclusive combination of schemas; top bars count the quantity kinds grounded in exactly that combination, and left bars give each schema's total distinct kinds. Six quantity kinds form a shared core common to all four schemas, while others are specific to individual schemas or perspective-based subsets.}
    \label{fig: quantityKing-upset}
\end{figure}

\subsection{Cross-Schema Comparison}
\label{sec:cross-schema}

Having described the four schemas individually, we now examine them collectively. Although each schema targets a distinct process (deposition or etching) and perspective (experimental or simulation), the four schemas are not independent artefacts. They share a common structural and semantic core while diverging in ways that reflect the specific process and perspective each serves. This section examines that pattern at two levels: the semantic grounding of quantities, and the architectural organisation of concepts, drawing on the quantity-kind overlap in Figure~\ref{fig: quantityKing-upset} and the concept alignment in Table~\ref{tab:concept-alignment}. A complementary comparison of normalized domain properties is provided in Appendix~\ref{app:domain-property-overlap}.

At the QUDT grounding level, the four schemas share a substantial common vocabulary. Figure~\ref{fig: quantityKing-upset} presents this as an UpSet plot, which is a generalisation of the Venn diagram to more than a few sets, in which each column denotes an exclusive combination of schemas and the bar above it counts the QUDT quantity kinds grounded in exactly that combination. The largest intersection is a core of six quantity kinds common to all four schemas:  Energy, Length, Pressure, Temperature, Thickness, and Time, denoting the physical quantities that any ALD or ALE record must express. This core is partly a product of the curation process itself: harmonising related quantities to a single QUDT grounding, such as mapping every pressure to \texttt{StaticPressure} and every growth-per-cycle deposition to \texttt{Thickness}, moved quantities that were previously grounded inconsistently into the shared core.

The conceptual alignment in Table~\ref{tab:concept-alignment} shows a similar shared foundation. All four schemas represent the process chemistry, substrate or material being processed, process conditions, and resulting material properties, although these concepts occur under different structural paths and at different levels of granularity. The normalized domain-property analysis in Appendix~\ref{app:domain-property-overlap} identifies five properties shared across all four schemas: film chemical composition, pressure, primary precursor or reactant, reactant flow rate, and substrate material. The same analysis also shows stronger perspective-based overlap between the two simulation schemas, including shared concepts related to computational methods, surface chemistry, reactor conditions, and predicted film behaviour.

  
\begin{table}[!t]
\centering
\footnotesize
\setlength{\tabcolsep}{4pt}
\renewcommand{\arraystretch}{1.2}
\caption{Concept alignment across the four ALD/ALE schemas. Each row is a modeling concept; cells give the schema group or property that represents it, and ``--'' marks a concept a schema does not model. Rows are shaded by how many schemas share the concept: \colorbox{covUniversal}{universal} (all four), \colorbox{covPartial}{partially shared} (two or three), and \colorbox{covSpecific}{schema-specific} (one).}
\label{tab:concept-alignment}
\resizebox{\textwidth}{!}{%
\begin{tabular}{@{}l l l l l@{}}
\toprule
\textbf{Concept} & \textbf{ALD exp.} & \textbf{ALD sim.} & \textbf{ALE exp.} & \textbf{ALE sim.} \\
\midrule
\rowcolor{covPartial}   Process type                 & aldMethod[].method & --                   & processType           & processType \\
\rowcolor{covSpecific}  Directionality               & --               & --                     & directionality        & -- \\
\rowcolor{covUniversal} Reactants / precursors       & reactantSelection & materials             & reactantSelection     & chemicalPrecursors[] \\
\rowcolor{covUniversal} Process conditions           & processParameters & reactorConditions     & processDetails        & processParameters \\
\rowcolor{covPartial}   \quad Process temperature    & temperature      & --                     & substrateTemperature  & temperature \\
\rowcolor{covUniversal} \quad Pressure               & pressure         & pressure               & pressure              & pressure \\
\rowcolor{covUniversal} \quad Gas / reactant flow    & flowRates[]      & precursorFlow, carrierGasFlow & reactantFlow   & gasFlowRates \\
\rowcolor{covUniversal} Resulting material / film properties & materialProperties & filmProperties  & etchedMaterialProperties & filmProperties \\
\rowcolor{covPartial}   \quad Uniformity             & uniformity       & uniformity             & etchedMaterialProperties & -- \\
\rowcolor{covPartial}   \quad Deposited-film roughness & roughness      & roughness              & --                    & -- \\
\rowcolor{covPartial}   \quad Etched-surface roughness & --             & --                     & etchedMaterialProperties & filmProperties \\
\rowcolor{covPartial}   \quad Density                & filmDensity      & density                & --                    & filmProperties \\
\rowcolor{covSpecific}  Device properties            & deviceProperties & --                     & --                    & -- \\
\rowcolor{covSpecific}  Area-selective growth mechanism & --            & selectiveGrowthMechanism & --                  & -- \\
\rowcolor{covSpecific}  Etch selectivity             & --               & --                     & selectivity           & -- \\
\rowcolor{covSpecific}  Process window               & --               & --                     & aleWindow             & -- \\
\rowcolor{covSpecific}  Synergy                      & --               & --                     & synergy               & -- \\
\rowcolor{covPartial}   Simulation method            & --               & simulationParameters   & --                    & simulationMethodology \\
\bottomrule
\end{tabular}%
}
\end{table}

\section{Application \& Use Case}
The schemas are intended not only as specifications but also as targets for structured knowledge extraction. This role has been demonstrated for the ALD experimental schema in our prior work, using an agentic large-language-model workflow for extracting structured process records from scientific literature~\cite{sadruddin2026designing}. In this work, the schema acts as the extraction blueprint, where publications describing \textit{\ce{ZnO}} and \textit{\ce{IGZO}} ALD processes, drawn from the AtomicLimits database, are processed to populate the corresponding schema properties. The generated records are checked for schema conformity, and extracted chemical entities are normalized to \href{https://pubchem.ncbi.nlm.nih.gov/}{PubChem} identifiers. The schema therefore defines both the information to be extracted and the structure that the resulting output must follow, transforming unstructured scientific text into machine-readable process records suitable for downstream comparison and knowledge-graph integration.

This work illustrates several ways in which schema-guided extraction improves the reliability and interpretability of the resulting records. First, typed properties and the structured QUDT representation of quantitative values enable automatic validation of data types, required structures, quantity kinds, permitted units, and explicitly defined numerical constraints. This prevents structurally invalid or physically impossible values, such as negative durations or flow rates, from being silently accepted. Values that remain structurally valid but appear scientifically implausible can also be identified more readily during subsequent review. For example, a reported growth-per-cycle value of 115~\AA{} was traced and marked incorrect as a film-thickness increment. Second, applying the schema across a corpus provides an empirical view of reporting coverage. Chemistry-related properties, including the deposited material, precursors, co-reactants, and process gases, were populated relatively consistently, whereas quantitative properties such as growth per cycle, dosing duration, and purge duration were reported less uniformly. While this use case focuses on the ALD experimental schema, the same schema-guided workflow can in principle be applied to the remaining experimental and simulation schemas.

\section{Conclusion}
This work presented four QUDT-grounded JSON Schemas for experimental and simulation studies of atomic layer deposition and atomic layer etching. Developed using \textsc{schema-miner}, the schemas capture a shared process-level foundation while preserving process and perspective-specific concepts. QUDT grounding represents quantitative values explicitly through value, unit, and quantity kind, supporting unit-aware validation and cross-publication comparison. Their use in an agentic literature-extraction workflow further demonstrates their practical value as targets for transforming scientific articles into structured, schema-conformant records.

Current ontology alignment focuses on quantitative properties through QUDT. Future work will extend this grounding to non-quantitative concepts, including materials, processing methods, structures, properties, and characterization or simulation activities, using materials-science ontologies such as the Platform MaterialDigital Core Ontology (PMDco)~\cite{bayerlein2024pmd}. Together with further domain-expert refinement and evaluation on broader literature corpora, this will strengthen cross-schema alignment and support more interoperable, reusable, and AI-ready ALD/ALE knowledge.

\begin{acknowledgments}
This work was supported by the “AI-Aware Pathways to Sustainable Semiconductor Process and Manufacturing Technologies (AWASES)” project, funded by Merck and Intel. The Large Language Models (LLMs) used during the experimentation were supported by KISSKI AI Service Center (BMBF, Grant ID: No. 01IS22093C).
\end{acknowledgments}

\section*{Declaration on Generative AI} 

During the preparation of this work, the author(s) used AI services in order to assist with refining manuscript text, and check grammar and spelling. After using this tool, the author(s) reviewed and edited the content as needed and take full responsibility for the publication's content.

\bibliography{sample-ceur}

\clearpage
\appendix

\section{Representative QUDT-Grounded Property Structure}
\label{app:qudt-property-example}

Figure~\ref{lst:plasmapower} shows the full QUDT-grounded representation of a quantitative property, using \texttt{plasmaPower} from the ALD experimental schema as an example. All quantitative properties across the four schemas follow this same \texttt{quantityValue}/\texttt{quantityKind} structure.

\begin{figure}[!htbp]
\caption{QUDT-grounded representation of the \texttt{plasmaPower} property in the ALD-experimental schema, illustrating the \texttt{quantityValue}/\texttt{quantityKind} structure used for all grounded quantities. The prefixes \texttt{qudt:} and \texttt{unit:} abbreviate \url{http://qudt.org/vocab/quantitykind/} and \url{http://qudt.org/vocab/unit/}, respectively.}%
\label{lst:plasmapower}%
\begin{lstlisting}[style=jsonstyle]
{
  "plasmaPower": {
    "type": "object",
    "description": "Plasma power (in Watts) if PEALD was used.",
    "required": ["quantityValue", "quantityKind"],
    "additionalProperties": false,
    "properties": {
      "quantityValue": {
        "type": "object",
        "required": ["numericValue", "unit"],
        "properties": {
          "numericValue": { "type": "number", "minimum": 0 },
          "unit": {
            "type": "object",
            "required": ["hasQuantityKind", "sameAs"],
            "properties": {
              "hasQuantityKind": { "const": "qudt:Power" },
              "sameAs": { "enum": ["unit:W", "unit:MilliW"] }
            }
          }
        }
      },
      "quantityKind": { "const": "qudt:Power" }
    }
  }
}
\end{lstlisting}
\end{figure}

\section{ALE Experimental and Simulation Schema Structures}
\label{app:ale-schema-structures}

This appendix presents the complete hierarchical structure of the two ALE schemas, complementing the ALD experimental and simulation schema diagrams in Section~\ref{fig:ald-exp-tree}. The diagrams follow the same conventions used throughout: each node shows a property name and, where applicable, its type. An asterisk (\textsuperscript{*}) marks a property required for the process, a dagger (\textsuperscript{\dag}) marks a QUDT-grounded quantitative property, whose internal \texttt{quantityValue}/\texttt{quantityKind} structure is not expanded (see Figure~\ref{lst:plasmapower}), and a double dagger (\textsuperscript{\ddag}) marks a property group whose further nested structure is omitted for simplicity. Full definitions for all properties are available in the published schemas in the Zenodo repository.

\clearpage
\subsection{ALE Experimental Schema}
\label{app:ale-experimental-schema}

\begin{figure}[!htbp]
\centering
\begin{adjustbox}{max width=\textwidth}
\begin{forest}
for tree={
  grow'=0,
  parent anchor=east,
  child anchor=west,
  anchor=west,
  edge path={
    \noexpand\path[\forestoption{edge}]
      (!u.parent anchor) -- +(5pt,0) |- (.child anchor)\forestoption{edge label};
  },
  edge={thin},
  l sep=10pt,
  s sep=2pt,
  font=\footnotesize,
}
[ALE experimental schema
  [etchedMaterial$^{*}$ : string]
  [processType$^{*}$ : string]
  [{directionality$^{*}$ :
        enum\{anisotropic, isotropic\}}]
  [reactantSelection$^{*}$
    [reactantA$^{*}$ : string]
    [reactantB$^{*}$ : string]
  ]
  [processDetails$^{*}$
    [substrateTemperature : QUDT quantity$^\dagger$]
    [pressure : QUDT quantity$^\dagger$]
    [substrate : string]
    [pulseTimes : object$^\ddagger$]
    [purgeTimes : object$^\ddagger$]
    [reactorType : string]
    [plasmaPower : QUDT quantity$^\dagger$]
    [rfPower : QUDT quantity$^\dagger$]
    [reactantFlow : QUDT quantity$^\dagger$]
    [numberOfCycles : QUDT quantity$^\dagger$]
  ]
  [etchControl$^{*}$
    [etchPerCycle$^{*}$ : QUDT quantity$^\dagger$]
    [massLoss : QUDT quantity$^\dagger$]
  ]
  [synergy$^{*}$
    [synergyValue : QUDT quantity$^\dagger$]
  ]
  [aleWindow$^{*}$
    [temperatureWindow : object$^\ddagger$]
    [ionEnergyWindow : object$^\ddagger$]
  ]
  [etchedMaterialProperties$^{*}$
    [materialProperties : object$^\ddagger$]
  ]
  [selectivity$^{*}$
    [selectivityDescription : string]
  ]
  [otherAspects$^{*}$
    [safety : string]
    [sustainability : string]
    [environmentalImpact : string]
  ]
]
\end{forest}
\end{adjustbox}

\vspace{2pt}
{\footnotesize $^{*}$ Required within its containing object;\quad $^\dagger$ QUDT-grounded quantitative property;\quad $^\ddagger$ additional nested structure omitted.}

\caption{Overview of the ALE experimental schema. The figure shows all top-level properties and the immediate children of each top-level property group and deeper nested structures are omitted where indicated. The scalar top-level properties \texttt{etchedMaterial}, \texttt{processType}, and \texttt{directionality} describe the etching target and overall process classification, while the remaining top-level properties organize reactants, process conditions, etch-control metrics, operating windows, material properties, selectivity, and related aspects.}
\label{fig:ale-exp-tree}
\end{figure}

\clearpage
\subsection{ALE Simulation Schema}
\label{app:ale-simulation-schema}

\begin{figure}[!htbp]
\centering
\begin{adjustbox}{max width=\textwidth}
\begin{forest}
for tree={
  grow'=0,
  parent anchor=east,
  child anchor=west,
  anchor=west,
  edge path={
    \noexpand\path[\forestoption{edge}]
      (!u.parent anchor) -- +(5pt,0) |- (.child anchor)
      \forestoption{edge label};
  },
  edge={thin},
  l sep=10pt,
  s sep=2pt,
  font=\footnotesize,
}
[ALE simulation schema
  [{processType$^{*}$ : enum\{thermal, plasma\}}]
  [chemicalPrecursors$^{*}$ : array<string>]
  [energySource$^{*}$ : string]
  [substrateMaterial$^{*}$
    [type$^{*}$ : string]
    [chemicalComposition$^{*}$ : string]
  ]
  [surfaceModel$^{*}$
    [dimensions$^{*}$ : object$^\ddagger$]
    [surfaceArea$^{*}$ : QUDT quantity$^\dagger$]
    [surfaceProperties$^{*}$ : object$^\ddagger$]
    [latticeParameters$^{*}$ : object$^\ddagger$]
  ]
  [processParameters$^{*}$
    [temperature$^{*}$ : QUDT quantity$^\dagger$]
    [pressure$^{*}$ : QUDT quantity$^\dagger$]
    [gasFlowRates$^{*}$ : object$^\ddagger$]
    [plasmaPower : QUDT quantity$^\dagger$]
    [exposureTime$^{*}$ : QUDT quantity$^\dagger$]
  ]
  [simulationResults$^{*}$
    [etchedMaterial$^{*}$ : string]
    [byProducts$^{*}$ : array<string>]
    [surfaceModifications$^{*}$ : string]
    [decompositionMechanisms$^{*}$ : string]
    [excitationDynamics : string]
    [etchRate$^{*}$ : QUDT quantity$^\dagger$]
    [surfaceDesorptionRates$^{*}$ : QUDT quantity$^\dagger$]
    [bindingAffinity$^{*}$ : string]
    [filmProperties : object$^\ddagger$]
    [designVariables : object$^\ddagger$]
    [fluorinationDetails : object$^\ddagger$]
    [selfLimitingReaction : object$^\ddagger$]
    [inertGasIrradiation : object$^\ddagger$]
  ]
  [simulationMethodology$^{*}$
    [methods$^{*}$ : array<string>]
    [methodDetails : object$^\ddagger$]
    [source : object$^\ddagger$]
  ]
]
\end{forest}
\end{adjustbox}

\vspace{2pt}
{\footnotesize $^{*}$ Required within its containing object;\quad $^\dagger$ QUDT-grounded quantitative property;\quad $^\ddagger$ additional nested structure omitted.}

\caption{Overview of the ALE simulation schema. The figure shows all top-level properties and the immediate children of each top-level property group and deeper nested structures are omitted where indicated. The schema organizes the modeled substrate and surface, process parameters, predicted simulation results, and computational methodology.}
\label{fig:ale-sim-tree}
\end{figure}

\section{Domain-Property Overlap Across Schemas}
\label{app:domain-property-overlap}

This section complements the QUDT quantity-kind comparison presented before by examining the overlap of normalized domain properties across the four ALD/ALE schemas. The comparison considers domain properties while excluding recurring QUDT helper properties, including \texttt{quantityValue}, \texttt{quantityKind}, \texttt{numericValue}, \texttt{unit}, \texttt{hasQuantityKind}, and \texttt{sameAs}. Properties with equivalent meanings but different schema-specific names were manually reviewed and mapped to a common normalized label, whereas conceptually distinct properties were retained separately, resulting in a UpSet plot showing semantic overlap between the schemas.

\begin{figure}[!htbp]
    \centering
    \includegraphics[width=1\textwidth]{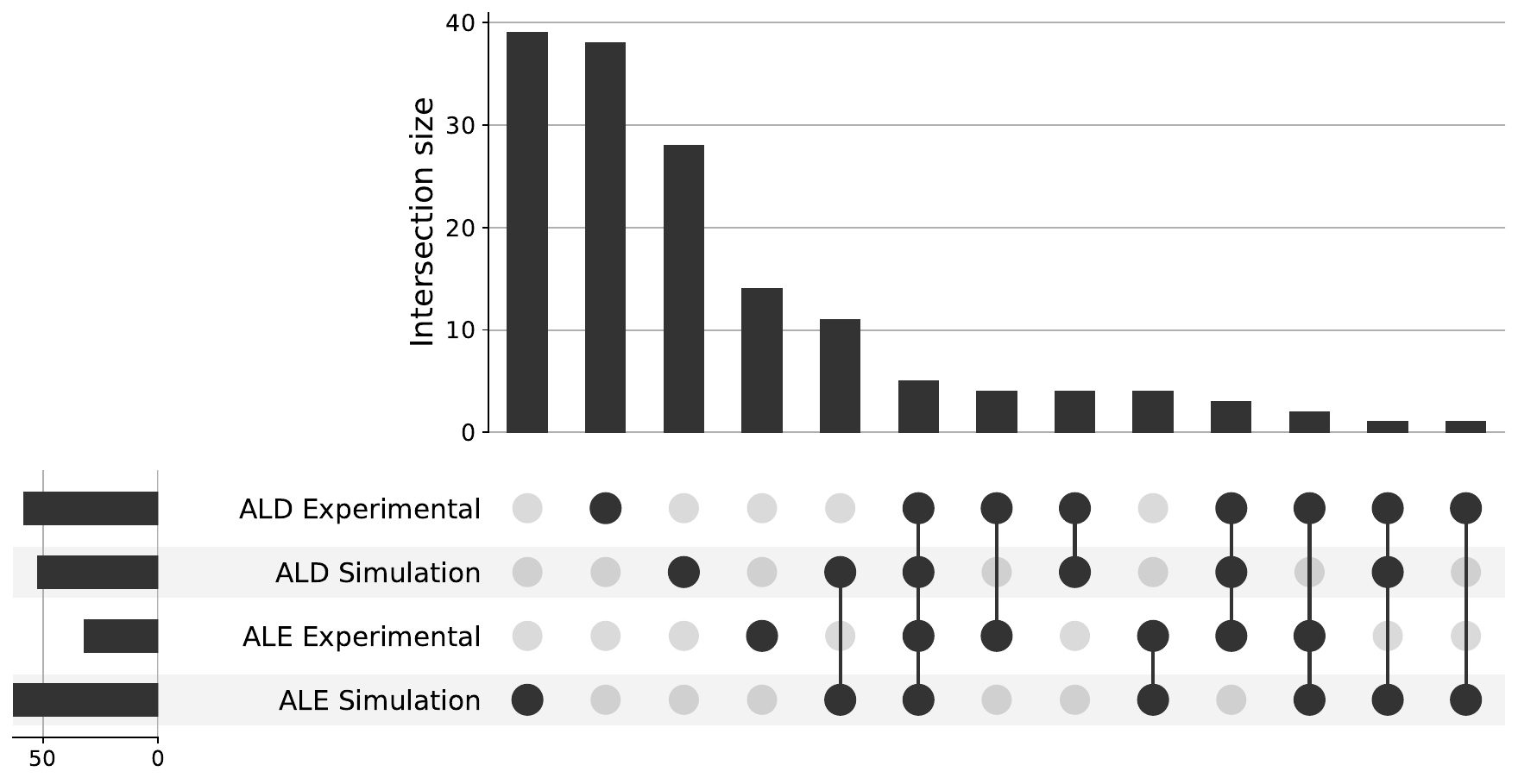}
    \caption{UpSet plot of normalized domain-property overlap across the four ALD/ALE schemas. Five properties form a shared core, while the remaining properties are schema-specific or shared by subsets of schemas.}
    \label{fig: quantityKing-upset}
\end{figure}

\end{document}